\documentclass[10pt,twocolumn,letterpaper]{article}

\usepackage{wacv}
\usepackage{times}
\usepackage{epsfig}
\usepackage{graphicx}
\usepackage{amsmath}
\usepackage{amssymb}
\usepackage{cite}
\usepackage{url}
\usepackage{multirow}
\usepackage{subcaption}
\usepackage{booktabs} 
\usepackage{bm}
\usepackage{framed}
\usepackage{soul}
\usepackage{comment}
\usepackage{latexsym}
\usepackage{epigraph}
\usepackage[pagebackref=true,breaklinks=true,letterpaper=true,colorlinks,bookmarks=false]{hyperref}

\usepackage[]{algorithm2e}
\usepackage{tabularx, booktabs} 
\newcolumntype{Y}{>{\centering\arraybackslash}X}
\usepackage{array}
\newcolumntype{M}{>{\centering\arraybackslash}m{9em}}
\newcolumntype{N}{>{\centering\arraybackslash}m{2.5em}}
\newcolumntype{P}{>{\centering\arraybackslash}m{4em}}
\newcolumntype{Q}{>{\centering\arraybackslash}m{10em}}
\newcolumntype{R}{>{\centering\arraybackslash}m{12.5em}}
\newcolumntype{S}{>{\centering\arraybackslash}m{5em}}
\usepackage{enumitem}
\usepackage{subcaption}
\graphicspath{{fig/}}

\DeclareMathOperator*{\argmax}{arg\,max}

\wacvfinalcopy 

\def\wacvPaperID{482} 

\ifwacvfinal\fi
\begin{document}

\setlength{\abovedisplayskip}{3pt}
\setlength{\belowdisplayskip}{3pt}

\title{
Toward Explainable Fashion Recommendation
}

\author{Pongsate Tangseng\hspace{2cm} Takayuki Okatani \\
Tohoku University\\
{\tt\small \{tangseng,okatani\}@vision.is.tohoku.ac.jp}
}


\maketitle

\begin{abstract}
Many studies have been conducted so far to build systems for 
recommending fashion items and outfits. Although they achieve good performances in their respective tasks, most of them cannot explain their judgments to the users, which
compromises their usefulness. Toward explainable fashion recommendation, this study proposes a system that is able not only to provide a goodness score for an outfit but also to explain the score by providing reason behind it. For this purpose, we propose a method for quantifying how influential each feature of each item is to the score.  Using this influence value, we can identify which item and what feature make the outfit good or bad. We represent the image of each item with a combination of human-interpretable features, and thereby the identification of the most influential item-feature pair gives useful explanation of the output score. To evaluate the performance of this approach, we design an experiment that can be performed without human annotation; we replace a single item-feature pair in an outfit so that the score will decrease, and then we test if the proposed method can detect the replaced
item-feature pair correctly using the above influence values. The experimental results show that the proposed method can accurately detect bad items in outfits lowering their scores. 


\end{abstract}

\section{Introduction}


Recently, there have been many studies of applying computer vision techniques to various problems of fashion, such as quantifying/measuring goodness of outfits~\cite{fashion144k, han2017learning, mining_outfit, tangsengwacv18, feng2018interpretable} and recommending to users
outfits from a pool of items~\cite{magic_closet, tangsengwacv18} or
outfits that fit users's personal  preferences~\cite{collaborative_fashion_recommendation} or location~\cite{zhang2017trip}. However, many of the existing studies, particularly the recent ones that employ CNNs, rely on black-box models, which may provide good performance on respective tasks but cannot explain the reason of their judgments
~\cite{collaborative_fashion_recommendation, magic_closet, mining_outfit, tangsengwacv18}. There are a few attempts to develop models that can provide useful explanations~\cite{zhang2017trip, feng2018interpretable}, but they require a large amount of manually annotated data for supervised training of the models, which is expensive and usually not publicly available. 

\begin{figure}[t]
\centering
  \includegraphics[width=.8\columnwidth]{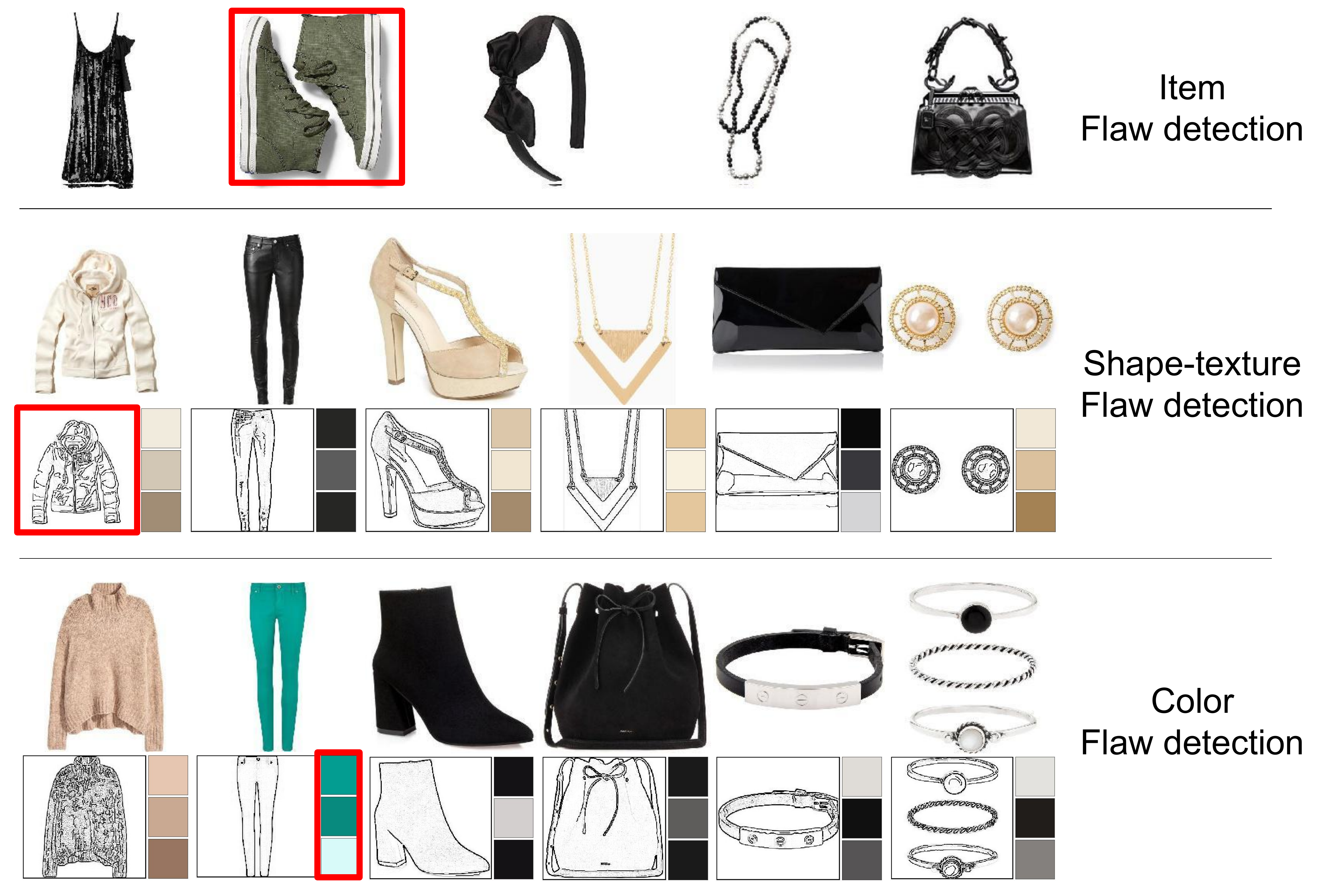}
  \caption{Our system first predicts a goodness score of an input outfit consisting of multiple items. It then identifies which item and what feature is the cause of, for instance, a low score. It is able not only to perform item-level identification (first row) but also to perform feature-level identification (second and third rows). }
  \label{fig:paper_goal}
\end{figure}

In this study, we propose a system that is able not only to judge and quantify goodness/badness of an outfit but also to provide a reason(s) of the prediction. Similar to existing methods, our system receives images of multiple items comprising an outfit as inputs and then computes a score quantifying its goodness/badness of the outfit; example inputs are shown in the rows of Fig.~\ref{fig:paper_goal}.  This forward computation is done by a part of our system called the outfit grader. To explain the output score, we quantify and use {\em how large the influence of each item, or of each feature of each item, is on the predicted score.} This enables to identify which item and what feature make the outfit good or bad; examples of the identification are shown in Fig.~\ref{fig:paper_goal}. For this purpose, we represent each item, rigorously its image, with a combination of human-interpretable features, and thereby the identification of the most influential item-feature pair will be a useful explanation of the score. 

To measure the influence of item-feature pairs, we employ the multiplication of an individual feature with the gradient of the output score with respect to the feature. This is similar to the methods for visualizing inference of CNNs, such as the multiplication of an input image with its sensitivity map \cite{szegedy2013intriguing,smilkov2017smoothgrad} (i.e., the score gradient with respect to image pixels) and Grad-CAM \cite{selvaraju2017grad}. The values thus computed are averaged and normalized within each feature of each item to yield our measure of the influence of the item-feature pair, which we call its \textit{Item-Feature Influence Value} (IFIV). Note that our method does not need extra training data other than those for training the outfit grader.  


It is usually hard to evaluate explanations provided by AI systems, since their quality can theoretically be evaluated only by humans. Human evaluation is generally costly; moreover, in our case, it is difficult to perform and conveys open problems, as the judgments to be explained are often subjective. To cope with this difficulty, we employ an automatic evaluation method by designing a test for the evaluation that is based on synthesis of datasets. The basic idea is that i) we first replace a single item or its single feature of an outfit so that the resulting score will decrease and ii) we then test if the proposed method can detect the replaced item by identifying the item-feature pair with the maximum IFIV. 




The organization of this paper is as follows. We first discuss the related work in Sec.~\ref{sec:ch4_related_work}. Next, we describe the proposed method for explaining judgments made by our outfit grader on the quality of input outfits in Sec.~\ref{sec:explaining_outfit_fashionability}. Section~\ref{sec:outfit_grader_evaluation} explains and evaluates the outfit grader that is the target of explanation. Experimental results on the proposed method for explaining its judgments are provided in 
Sec.~\ref{sec:outfit_flaw_detection}. Section~\ref{sec:ch4_conclusion} concludes this study. 

\section{Related Work}\label{sec:ch4_related_work}

\subsection{Measuring Goodness of Outfits}\label{subsec:outfit_grader_recommendation}

There is a growing interest in the application of computer vision techniques to measure the goodness of outfits. The authors of~\cite{fashion144k} predicted fashionability scores from an outfit image and tags. 
The authors of~\cite{han2017learning} use bidirectional LSTM (Bi-LSTM)~\cite{graves2012supervised} to learn the compatibility relationship among fashion items by modeling an outfit as a sequence, whereas fully-connected layers are employed in \cite{mining_outfit, tangsengwacv18}. In~\cite{han2017learning, mining_outfit, tangsengwacv18}, CNNs trained for generic image recognition are used to extract features for their respective purposes. Overall, the proposed methods in these studies work fairly well for measuring the goodness of outfits, i.e., predicting a score for each outfit. However, these methods lack the ability of providing reasons of the predicted scores. 

\subsection{Explaining Inference of Models}\label{subsec:XAI}
Recent advances in deep learning have dramatically improved accuracy of many computer vision tasks, such as image classification~\cite{VGG, resnet, inceptionv3}, object detection~\cite{fasterRCNN}, object segmentation~\cite{FCN, mnc}, Visual-Question Answering (VQA)~\cite{antol2015vqa, gao2015you, malinowski2015ask, ren2015exploring}, etc. These progresses have left behind explanation and understanding of what the deep neural networks have learned  
as well as how they make inference/judgments. Thus, there is a growing concern particularly about life-critical applications~\cite{lipton2016mythos}. A number of studies have been conducted to resolve this so far; ~\cite{ribeiro2016should, zhou2016learning, selvaraju2017grad, bau2017network} to name a few. LIME ~\cite{ribeiro2016should} is a method for explaining the prediction of a machine learning model for an input, which estimates a linear model that locally approximates the model at the neighborhood of the input, and then uses it for explanation. There are many studies of visualization of inference made by CNNs. The authors of~\cite{zhou2016learning} proposed the Class Activation Map (CAM) for a particular class of CNN models, which shows the region in the input image that is responsible for the prediction. This is later extended to Grad-CAM~\cite{selvaraju2017grad}, which is be applicable to more general CNN models, including 
image captioning~\cite{chen2015microsoft, johnson2016densecap, vinyals2015show}, and Visual Question Answering (VQA)~\cite{antol2015vqa, gao2015you, malinowski2015ask, ren2015exploring}.

\subsection{Explainable Models for Fashion}
The aforementioned computer vision systems for fashion ~\cite{magic_closet, amazon_data, collaborative_fashion_recommendation, mining_outfit,StreetStyle2017} employ black-box models, too, which show fairly good performance for the respective tasks but lack ability of providing reason of inference/judgment. 
It is not straightforward to apply the above generic methods for explaining machine learning and deep learning models to these systems for fashion, because the problems are basically more complicated (e.g., multiple items contained in an outfit, stratified factors affecting the goodness/badness of an outfit etc.)

There are 
a few studies that attempt to provide useful explanation on model's evaluation of outfits 
~\cite{lin2018explainable, feng2018interpretable}. The method proposed in \cite{feng2018interpretable} relies on a massive amount of annotated data to train a multi-category attribute predictor and create a composition graph based on pairwise co-occurrence of those predicted attributes in outfits. 
On the other hand, the method proposed in ~\cite{lin2018explainable} provides an upper-lower matching recommendation with textual explanation by utilizing comments provided by users of~\url{polyvore.com}. Although this method does not require manual annotation, it can deal with only two items in each outfit. 



\begin{figure*}[t]
\centering
  \includegraphics[width=.9\textwidth]{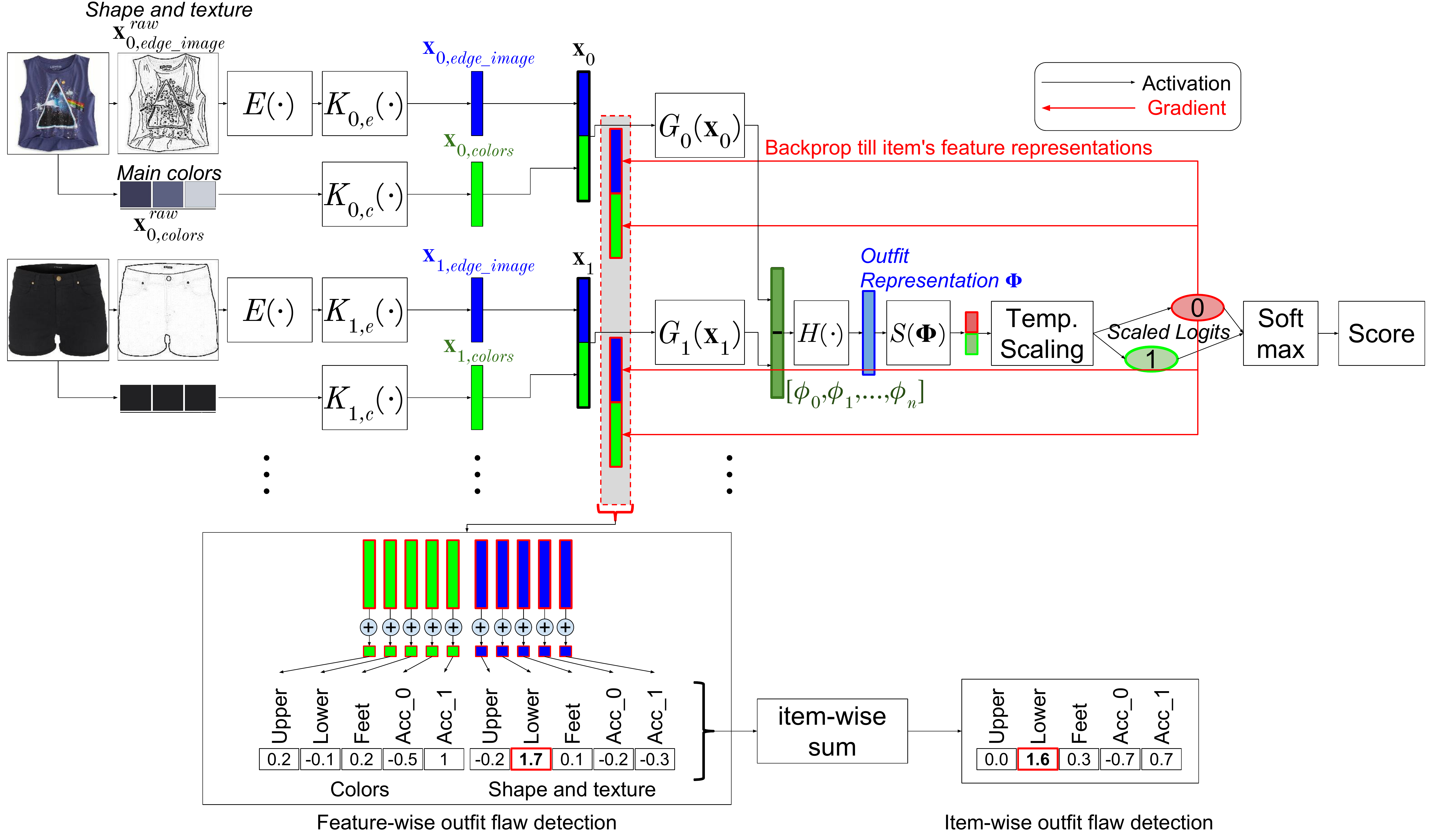}
  \caption{The overview of the proposed system. Given an outfit as a set of items, it extracts \textit{edge\_image} and main colors of each item. The \textit{edge\_image} is forward-propagated through a pretrained CNN $E$, then the output and main colors are forward-propagated through a series of concatenation and fully connected layers with ReLU (i.e., $K$, $G$ and $H$) to obtain the score. 
  The system also computes the gradient of the score (rigorously, the logit before softmax)
  with respect to the representation of each item through backpropagation. 
  The gradients are multiplied with the corresponding features, yielding \textit{Item Feature Influence Value} (IFIV).
  There is a single IFIV for each item-feature pair. } 
  \label{fig:branches_outfit_grader}
\end{figure*}

\section{Explaining Goodness of Outfit}\label{sec:explaining_outfit_fashionability}

Figure~\ref{fig:branches_outfit_grader} shows an overview of the proposed system.
It employs the outfit grader developed in \cite{tangsengwacv18}, which classifies an input outfit either as positive (a good outfit) or negative (a bad outfit). 
We wish to explain judgment made by the grader for an outfit, i.e., why it classifies an input outfit as positive or as negative.
For this purpose, we evaluate influence of each item and its features on the grader's judgment.
The former (i.e., the influence of each item) provides item-level explanations, e.g., {\em this outfit is bad because of the inclusion of this particular item}. For this, we use the internal features (i.e., penultimate layer activation) that the grader uses. To further enable to obtain deeper explanations, we use human-interpretable features for the purpose, e.g., shape, texture, and colors extracted from the item images comprising the input outfit. To do this, we redesign the grader so that it can make judgments solely from these features. 



\subsection{Interpretable Item Features} \label{subsec:edge_imge_colors_extraction}

The idea is to represent each item in terms of its {\em attributes} that are human-interpretable.  We also rebuild the grader so that it can judge an input outfit from its attribute representation, and then attempt to explain its judgments according to influence of each attribute on the final score. 

There are many candidate for this purpose, such as item type, brand, color, shape, texture, style etc. However, it may be a difficult task even for fashion experts to define such attributes determining the goodness of outfit. Moreover, we also need to be able to accurately predict those attributes from input item images, which will require costly annotation for training a proper model (e.g., a CNN). Additionally, the attributes need to be sufficiently rich so that  the grader can properly judge goodness of outfits only from them. 

Considering these requirements,  {\color{black} we choose primitive image features} that can be easily extracted from the item images: shape, texture, and colors. 
\begin{figure}[t]
\centering
  \includegraphics[width=.8\columnwidth]{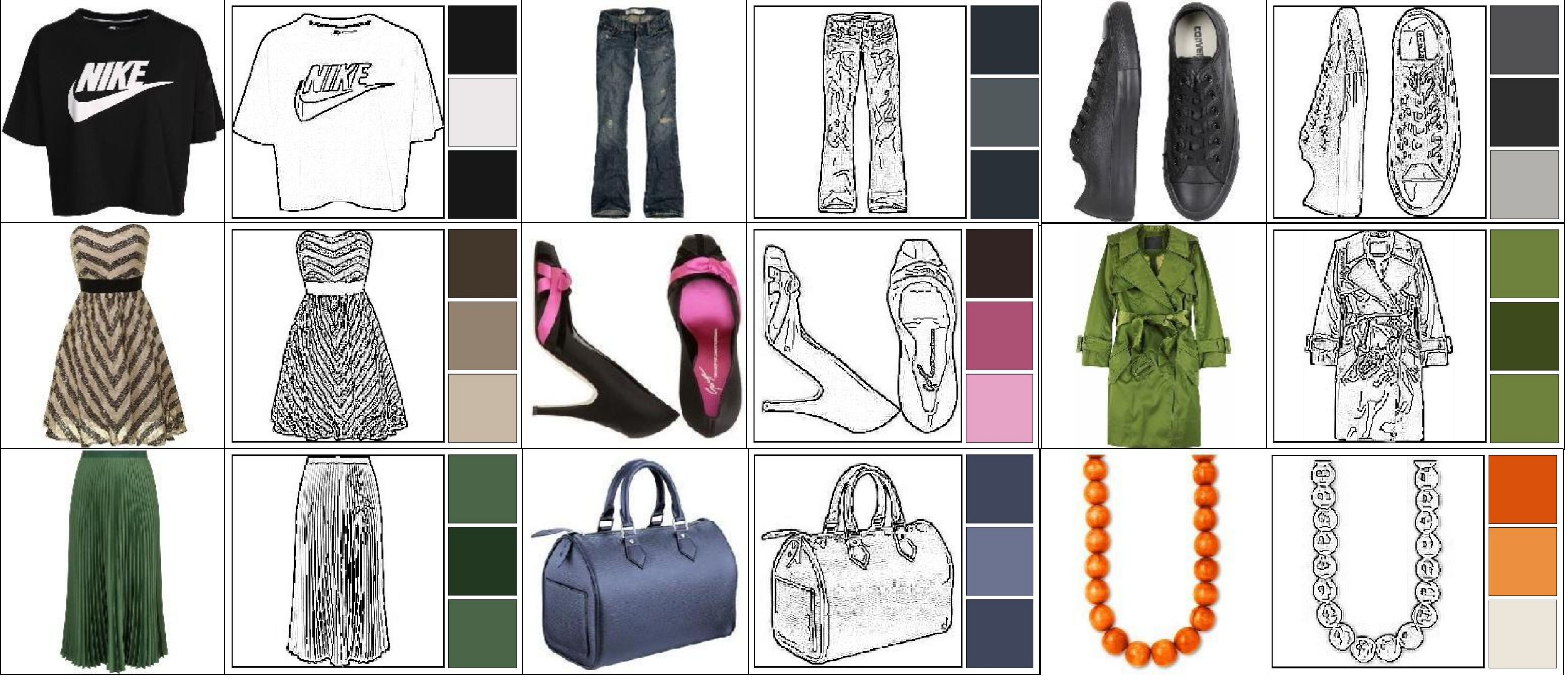}
  \caption{Item images with their \textit{edge\_image} and main three colors used as their features.} 
  \label{fig:image_edge_image_colors}
\end{figure}
To be specific, we first divide contents of item images into color and non-color information. For the former, we extract three dominant colors from each image by finding clusters of pixels in color space. For non-color information, we first convert the image into gray-scale and then extract edges, which are expected to maintain shape and texture of the item. Figure~\ref{fig:image_edge_image_colors} shows examples of original images, their \textit{edge\_image}, and three dominant colors. Their details are given below.


For colors, after removing background from the item image, we apply K-mean clustering~\cite{lloyd1982least} to cluster all the pixels in the item image into three main colors {\color{black} in RGB color space}. We use their centroids as three dominant colors of the item, yielding a 9-dimensional vector (3 colors $\times$ 3 RGB color values) for each item image.
We denote it by $\mathbf{x}^{raw}_{i,colors}$, where the subscript $i$ indicate that this is the color of the item that occupies $i$-th outfit part.
In addition, since we use a zero-vector to represent absence of an outfit part, to enable to deal with outfits with a variable number of items,
as in~\cite{tangsengwacv18}, we add 1 to all color values to avoid the conflict of a zero-vector with black color, resulting in the shift of the color value range from [0,1] to [1,2].

For shape and texture, we extract features in the following way. Let $I$ be the input item image.
We first apply the Canny edge detector \cite{canny1986computational} to $I$ to obtain an edge map $I_{e_1}$. In parallel, we also apply a simple $3\times 3$ filter $f$ to $I$ as $I_{e_2}= I \ast f$; $f$ is defined as
\begin{equation}
    f=
\begin{bmatrix}
    -1 & -1 & -1 \\
    -1 & 8 & -1 \\
    -1 & -1 & -1 \\
\end{bmatrix}.
\end{equation}
We add these two edge-like maps to obtain
\begin{equation}
    I_e=I_{e_1} + I_{e_2}.
\end{equation}
We call its black-white inverted version  (i.e., $I_e\leftarrow 255-clip(I_e, 0, 255)$) \textit{edge\_image} of $I$. We then use a pretrained convolutional neural network (CNN) to extract an $n$-dimensional embedding of \textit{edge\_image}, which we denote by 
$\mathbf{x}^{raw}_{i, edge\_image}$, as
\begin{align}\label{equ:item_encoder}
   \mathbf{x}^{raw}_{i, edge\_image} = E(edge\_image)
\end{align}
where $E$ is the CNN (up to its penultimate layer). We will  use 
this as the representation of shape and texture of 
the item occupying the $i$-th outfit part.
\label{subsec:item_representation}


The features $\mathbf{x}^{raw}_{i,colors}$ and $\mathbf{x}^{raw}_{i, edge\_image}$ obtained as above are transformed by a trainable item-feature encoders $K_{i,c}$ and $K_{i,e}$ into item-feature encodings $\mathbf{x}_{i,colors}$ and $\mathbf{x}_{i, edge\_image}$ respectively. We use a stack of a few fully-connected layers for $K_{i,c}$ and $K_{i,e}$ each.
Finally, we concatenate 
them together and denote the resultant vector by $\mathbf{x}_i= [\mathbf{x}_{i, edge\_image}^\top,\mathbf{x}_{i,colors}^\top]^\top$, 
which gives a representation of an item. 


\subsection{Outfit Grader}

Our outfit grader is basically the same as the one proposed in \cite{tangsengwacv18} except the representation of items described above. We summarize its design here. The input is an outfit consisting of $n$ items, each of which occupies a different part. 
Given the feature of an $i$-th part item as mentioned above, our grader first transforms it by a trainable item encoder $G_i$ as 
\begin{align}
    \phi_i = G_i(\mathbf{x}_i).
    \label{equation:item_encoder}
\end{align}
We use a stack of a few fully-connected layers for $G_i$.
The representations of $n$ items are then concatenated and transformed to the representation $\mathbf{\Phi}$ of the entire outfit as
\begin{align}
    \mathbf{\Phi} = H([\phi_0, \phi_1, \dots, \phi_n]),
\end{align}
where $H$ is a trainable outfit encoder, for which we employ a single fully-connected layer (followed by BN and ReLU). 

The grader performs binary classification on  the representation $\mathbf{\Phi}$ of the input outfit $O$. 
To do this, the outfit representation is transformed by a single fully-connected layer $S$ to two logits $\mathbf{s}=[s_{pos},s_{neg}]$ as $\mathbf{s}=S(\mathbf{\Phi})$. Then they are normalized by softmax to yield scores for positive and negative classifications. Denoting the score for $O$ being positive by $F(O)$, it is given by
\begin{equation}
    F(O)=\sigma_{pos}(\mathbf{s})=\frac{\exp{(s_{pos})}}{\exp{(s_{pos})}+\exp{(s_{neg})}}.
    \label{eqn:softmax}
\end{equation}


For the CNNs extracting item features (e.g., $\mathbf{x}_{edge\_image}$), we use those pretrained on other tasks such as object recognition. Thus, the learnable parameters in the grader are in $K_{i,e}$, $K_{i,c}$, $G_i$, $H$, and $S$. They are learned by minimizing a cross-entropy loss on training data consisting of pairs of outfit $O$ and the ground-truth  label (i.e., positive or negative).

\paragraph{Calibration of Outfit Scores}
\label{paragraph:temperature_scaling}

It is known \cite{guo2017calibration} that modern deep neural networks employing softmax for multi-class classification tend to be over-confindent, that is, the score of the predicted class, or {\em confident} (i.e., the max of softmax outputs), tends to be large and even close to one, even if the prediction is wrong. We found that this is exactly the case with our implementation of the outfit grader \cite{tangsengwacv18}. A simple but effective method to alleviate this overconfidence is to perform calibration of the softmax outputs using temperature scaling \cite{guo2017calibration,platt1999probabilistic}. To be specific, we replace $\mathbf{s}$ in the softmax (\ref{eqn:softmax}) with $\mathbf{s}/T$. 
$T$ is determined using validation samples so that the resulting score $F(O)$ is as close to classification accuracy as possible; then the score will better represent confidence of the prediction.  
We use $\hat{q} = 100 \cdot F(O)$ (in percent) as the fashionability score of an outfit $O$.


\subsection{Item Feature Influence Value (IFIV)}

Suppose that we input an outfit to the above grader and receive its judgment. To explain the judgment, we evaluate influence of each feature of each item. If the judgment is negative and a particular feature of an item has large influence on it, we regard that feature of the item to be the reason for the negativity; the same is true for a positive judgment. 

To be specific, we define the influence on the logit $s_c$ ($c \in \{neg,pos\}$) of a feature $f(\in\{edge\_image,colors\})$  of $i$-th item, denoted by $\mathbf{x}_{i,f}$, as follows. We first compute
\begin{equation}
    \mathbf{g}_{i,f}= \mathbf{x}_{i,f} \odot \frac{\partial s_c}{\partial \mathbf{x}_{i,f}},
\end{equation}
where $\odot$ is element-wise multiplication. Note that the logit $s_c$ here is the temperature-scaled version mentioned above. A similar method  is used for visualization of CNNs for object classification, where the pixel-wise multiplication of an input image and the gradient of a class score with respect to its pixels is used to show which part positively or negatively affects the score and which part has no influence on it. 
As we consider influence of only each feature, not its element, we compute the sum over  all its elements as
\begin{subequations}
 \label{equation:IFIV}
    \begin{equation}
    IFIV_{i} = \sum_f IFIV_{i,f},
    \end{equation}
    where
    \begin{equation}
    IFIV_{i,f} = \sum_k g_{i,f,k},
    \end{equation}
\end{subequations}
where $g_{i,f,k}$ is the $k$-th element of $\mathbf{g}_{i,f}$. 
Figure~\ref{fig:branches_outfit_grader} shows the diagram explaining how \textit{Item Feature Influence Value (IFIV)} of each item feature is computed. 

\section{Evaluation of the Outfit Grader}
\label{sec:outfit_grader_evaluation}

\subsection{Prediction Accuracy vs. Interpretability} \label{subsec:outfit_grader_configuration}

We redesign the outfit grader for the purpose of improved explanability. The original model \cite{tangsengwacv18} is designed to be an end-to-end model receiving raw item images as inputs, aiming at the best prediction accuracy of outfit quality. Our redesigned model receives hand-engineered features extracted from item images for the sake of explanability. This will sacrifice accuracy of outfit quality prediction. 
We conducted experiments to examine this. 

\paragraph{Model architecture}
We compare two models that differ only in the item representation $\mathbf{x}$. One is the model we described in Sec.~\ref{sec:explaining_outfit_fashionability}. The other is a baseline model, which uses a CNN feature directly extracted from RGB item images; to be specific, the feature of 
the $i$-th part item is given by $\mathbf{x}_i=E(RGB\_image)$, where $E$ is a pretrained CNN that is the same as the one used to extract 
$\mathbf{x}^{raw}_{i, edge\_image}$.
The configurations and parameters that are shared by the two models are as follows:
\begin{itemize}
\itemsep0em 
    \item For the feature extractor $E$, we employ ImageNet-pretrained InceptionV3~\cite{inceptionv3}. The activation of \textit{pool5} layer for an input item image is used for $\mathbf{x}$, which forms a  $2048$-dimensional vector.
    \item An identity function is used for 
    item-feature encoders $K_{i,e}$, $K_{i,c}$ and item encoder $G_i$.
    \item A single fully-connected layer with 4096 units is used for the outfit encoder $H$, followed by  batch normalization~\cite{ioffe2015batch} and ReLU~\cite{nair2010rectified} activation function.
    \item The both models are trained for 50 epochs with learning rate $1e-4$ and batch size 256 on Polyvore409k dataset~\cite{tangsengwacv18}.
\end{itemize}


\begin{table}[t]
\centering
\caption{Training, validation, and testing accuracy and average f1 of two outfit graders (a baseline and the  interpretable model) on Polyvore409k dataset~\cite{tangsengwacv18}.}
\label{table:RGB_vs_edge_image_colors}
\small\begin{tabular}{@{}cccc@{}} 
\toprule
\multirow{2}{*}{Partition} & \multirow{2}{*}{Metric} & \multicolumn{2}{c}{Model} \\ \cmidrule(l){3-4} 
 &  & Baseline & Interpretable \\ \midrule
\multirow{2}{*}{Train} & Acc. & 98.41 & 99.04 \\ 
 & Avg. F1 & 98.20 & 98.92 \\ \midrule
\multirow{2}{*}{Validation} & Acc. & 83.19 & 80.23 \\ 
 & Avg. F1 & 81.86 & 79.06 \\ \midrule
\multirow{2}{*}{Test} & Acc. & 79.19 & 76.36 \\ 
 & Avg. F1 & 74.11 & 71.42 \\ \bottomrule
\end{tabular}
\end{table}

\paragraph{Results}
Table~\ref{table:RGB_vs_edge_image_colors} shows the results. Accuracy indicates that of binary classification, where a prediction is considered to be correct if it matches the ground truth. As expected, the baseline model shows better performance than the interpretable model by 2.83\% accuracy and 2.69\% average f1. This is a noticeable gap but is arguably not so large to make the explanation by the interpretable model meaningless. 

\begin{figure}[t]
\centering
  \includegraphics[width=\columnwidth]{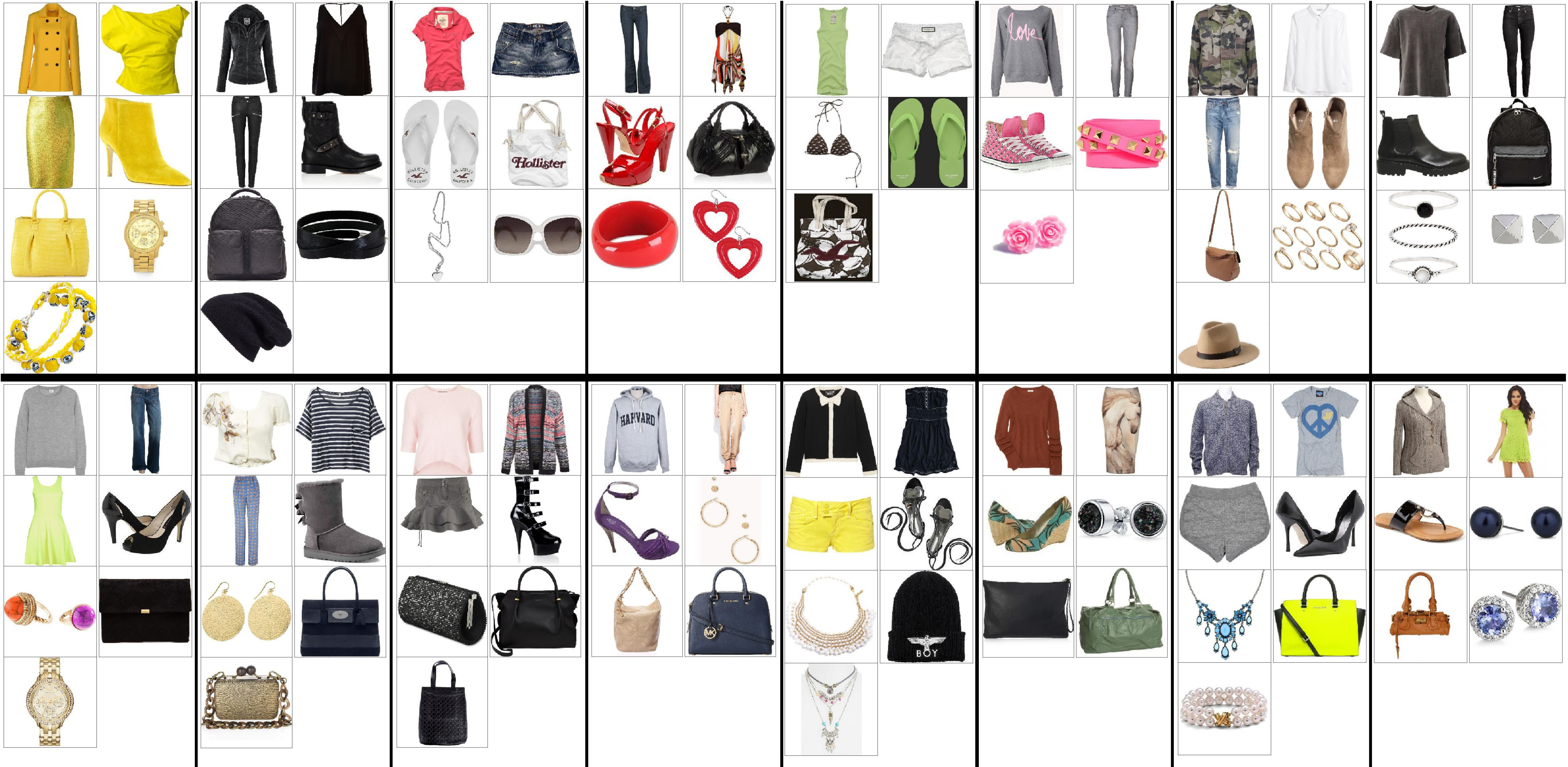}
  \vspace{-2em}
  \caption{The best (upper) and worst (lower) eight outfits from testing partition of Polyvore409k dataset according to our outfit grader.} 
  \label{fig:best_worst_outfits}
\end{figure}

\paragraph{Configuration of Outfit Grader}
To recover the performance drop as much as possible and further achieve better prediction accuracy, we tested a number of configurations of the interpretable grader. To be specific, we tested different configurations of the item-feature encoder $K_{i,c}$ and $K_{i,e}$, the item encoder $G_i$ and the outfit encoder $H$.
The configurations and their performance on testing samples are shown in Table~\ref{table:various_outfit_graders}. Since the model \#3 has the best performance, we will use this model for the experiments on explainability using feature influence values. 
Figure~\ref{fig:best_worst_outfits} shows examples of judgments of the grader; outfits with the highest score and those with the lowest scores. 
\begin{table}[t]
\centering
\caption{Testing accuracy and average f1 of various configurations of outfit grader after training for 50 epochs of Polyvore409k dataset~\cite{tangsengwacv18}. Each cell in the ``Item-feature Encoder $K_{i,c}$, $K_{i,e}$'', ``Item Encoder $G_i$'', and ``Outfit Encoder $H$'' columns specify the size of the fully-connected layer
The $\times$ indicates a stack of multiple layers.}
\vspace{-.5em}
\label{table:various_outfit_graders}
\small\begin{tabularx}{\columnwidth}{cSPPNN}
\toprule
\# & Item-feature Encoder $K_{i,c}$, $K_{i,e}$ & Item Encoder $G_i$ & Outfit Encoder $H$ & ~~~Acc.~~~ & Avg. F1 \\ \midrule
1 & - & - & 4096 & 76.36 & 71.42 \\
2 & 128 & 1024 & 2048 & 80.19 & 75.76 \\
\textbf{3} & \textbf{1024} & \textbf{1024} & \textbf{2048} & \textbf{80.75} & \textbf{76.76} \\
4 & 128 & 128 & 128 & 77.56 & 71.61 \\
5 & 128$\times$64 & 512$\times$256 & 2048 & 80.05 & 75.70 \\
6 & 128$\times$64$\times$32 & 512$\times$256 & 2048 & 79.04 & 75.84 \\ \bottomrule
\end{tabularx}
\end{table}

\subsection{Effect of Calibration of Score (Confidence)}
\begin{figure}[t]
\small
\centering
\begin{tabularx}{\columnwidth}{c *{2}{M}}
\toprule
  & Validation & Test\\
\midrule
\begin{tabular}[]{@{}c@{}}Before\\temperature\\scaling\end{tabular} & \includegraphics[width=.34\columnwidth]{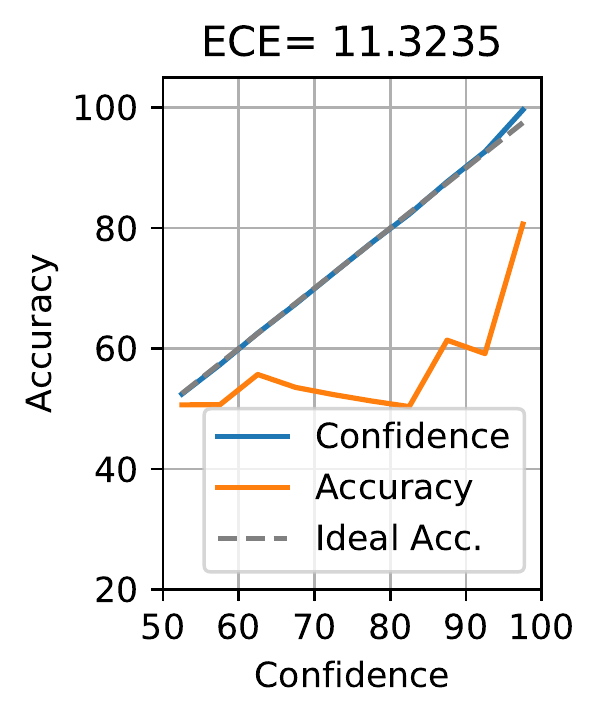} & 
\includegraphics[width=.34\columnwidth]{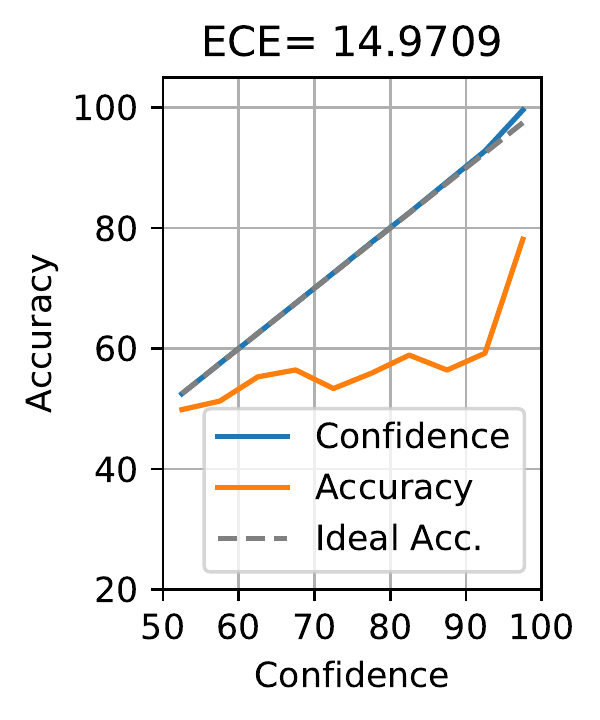} \\
\begin{tabular}[]{@{}c@{}}After\\temperature\\scaling\end{tabular} & \includegraphics[width=.34\columnwidth]{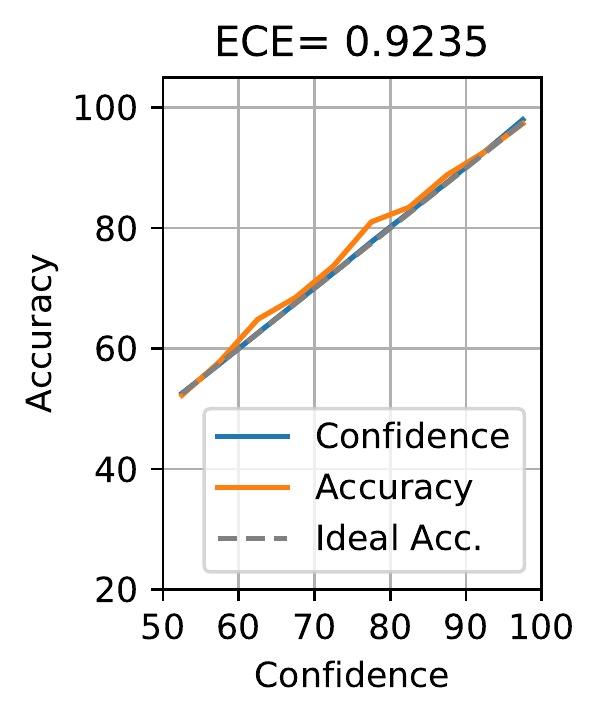} & 
\includegraphics[width=.34\columnwidth]{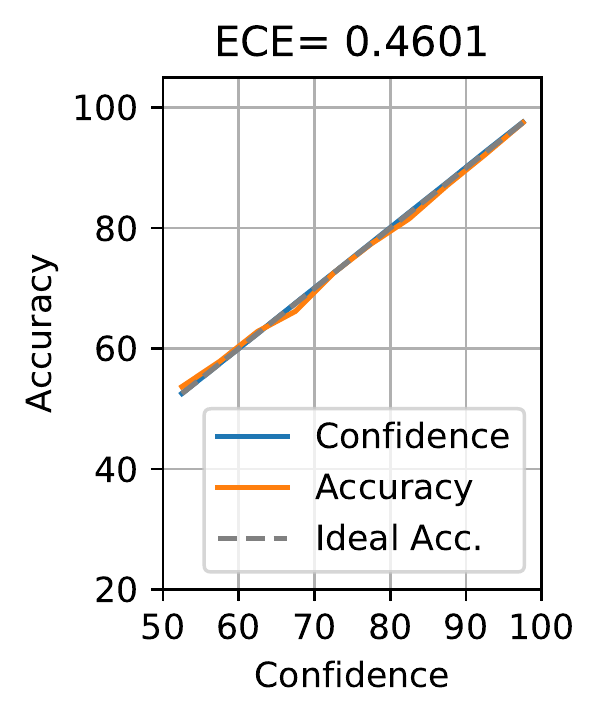} \\
\bottomrule
\end{tabularx}
\vspace{-1em}
\caption{Reliability diagrams and ECE values before and after temperature scaling for validation and testing partition of Polyvore409k dataset~\cite{tangsengwacv18}. {\color{black} Confidence is equivalent to the outfit score.}}
\label{fig:reliability_diagrams}
\end{figure}

\begin{figure}[t]
\small
\centering
\begin{tabularx}{\columnwidth}{c *{2}{M}}
\toprule
 & Validation & Test\\
\midrule
\begin{tabular}[]{@{}c@{}}Before\\temperature\\scaling\end{tabular} & \includegraphics[width=.34\columnwidth]{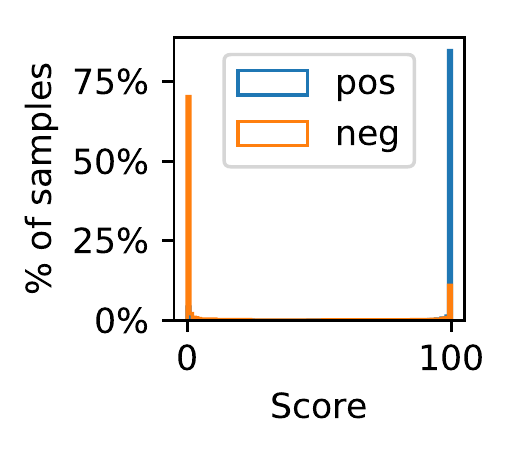} & 
\includegraphics[width=.34\columnwidth]{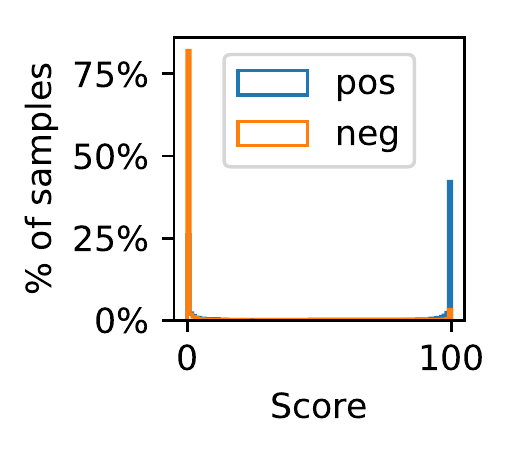} \\
\begin{tabular}[]{@{}c@{}}After\\temperature\\scaling\end{tabular} & \includegraphics[width=.34\columnwidth]{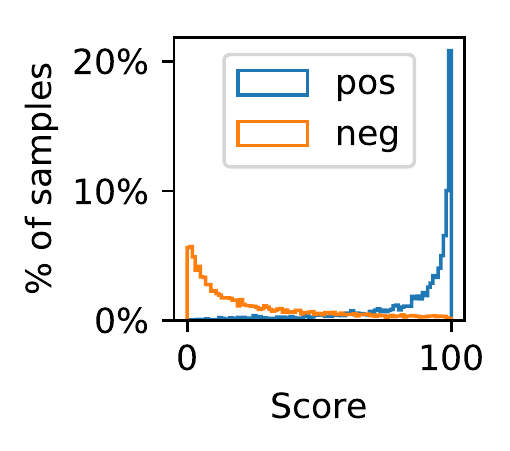} & 
\includegraphics[width=.34\columnwidth]{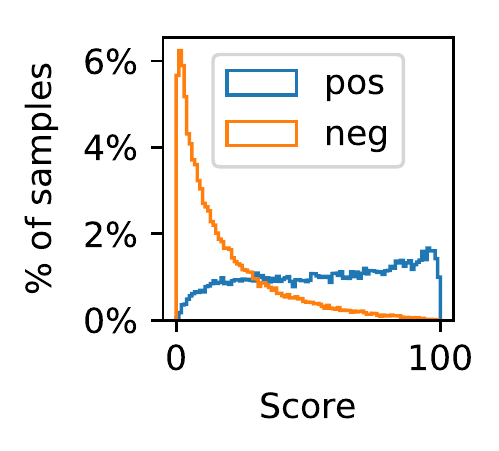} \\
\bottomrule
\end{tabularx}
\vspace{-1em}
\caption{Distribution of outfit scores before and after temperature scaling for positive and negative samples in validation and testing partition of Polyvore409k dataset~\cite{tangsengwacv18}.}
\label{fig:score_distributions}
\end{figure}

As mentioned in Sec.~\ref{paragraph:temperature_scaling}, we employ the temperature scaling to calibrate the outfit score (or confidence) $\hat{q}$. Figure~\ref{fig:reliability_diagrams} shows the reliability diagrams~\cite{degroot1983comparison, niculescu2005predicting} before and after the calibration.
Searching for the best value for the temparature $T$ on the validation samples yielded $T=6.77$. 
To do this, we split all the testing samples into 10 bins with an equal width, using which we plot the expected accuracy of samples in each bin against the average confidence from the outfit scores. 
A perfectly calibrated model will yield an identity relation between them. We also calculated expected calibration error (ECE)~\cite{naeini2015obtaining},  the difference in expectation between confidence and accuracy. 
ECE is reduced from 11.32 and 14.97 before the calibration to 0.92 and 0.46 after calibration for validation and testing partition of Polyvore409k dataset~\cite{tangsengwacv18} respectively. Figure \ref{fig:score_distributions} shows distributions of outfit scores for samples with positive labels and those with negative labels. The distributions with the temperature scaling clearly have a much wider spread, making the score more meaningful. We can conclude from Figs.~\ref{fig:reliability_diagrams} and \ref{fig:score_distributions} that the temperature scaling is able to calibrate the outfit scores.

\section{Experimental Results}
\label{sec:outfit_flaw_detection}

We conducted experiments to evaluate the proposed method for explaining judgment of the outfit grader. For the grader, we used the \textit{1024-1024-2048} outfit grader from Table~\ref{table:various_outfit_graders}.


\subsection{Experimental Design}


Suppose that an outfit is bad (i.e., not fashionable) due to a single item contained in it. There should also be a reason why the item does not match the outfit and makes it bad, e.g., because of its incompatible color or its unmatched shape and texture. We want to identify the item as well as the reason for the bad outfit. 

Based on the proposed framework, this is formulated as a task of  identifying the item-feature pair that has the most negative influence on an input outfit. We apply the proposed method to this task and evaluate its performance. 

\begin{figure}[t]
\centering
  \includegraphics[width=\columnwidth]{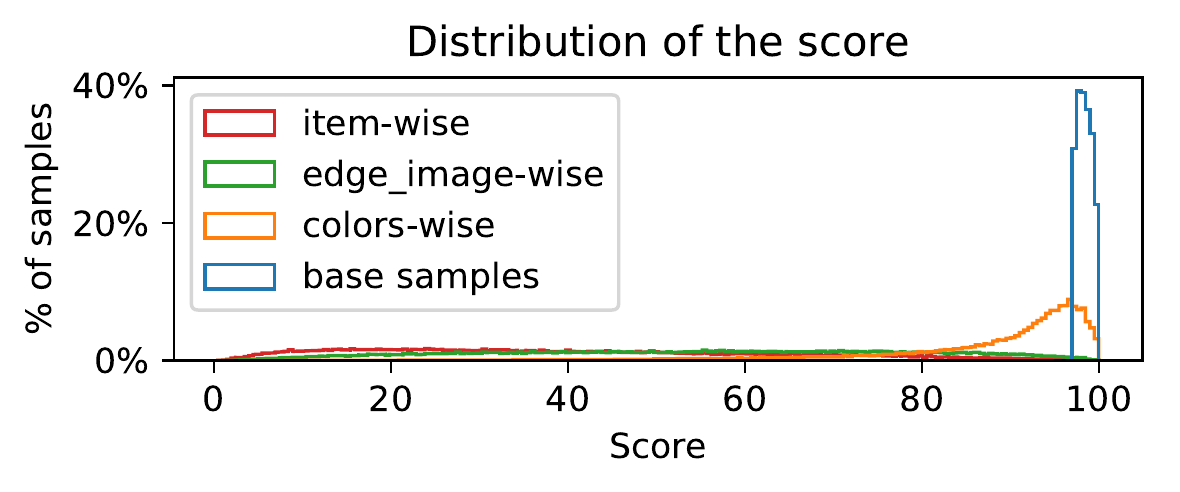}
  \vspace{-2.5em}
  \caption{The distribution of scores of each type of samples.} 
  \label{fig:sample_score_distribution}
  \vspace{-1em}
\end{figure}

For this purpose, we create a set of negative outfits from positive ones in the dataset in the following way. For a positive outfit, we choose an item from those contained in it and then replace its feature $f(\in \{edge\_image, colors\})$ 
and ensure that the replacement does decrease the outfit score. 
Note that we are interested here not in the correctness of the judgment of the outfit grader  but in how well its judgment can be explained, more precisely, accuracy of the proposed method identifying the item-feature pair lowering the score.  
Detailed procedures for the creation of data are as follows:
\begin{enumerate}
\itemsep0em 
    \item 1,000 base outfits with the highest scores are chosen from the test partition of Polyvore409k dataset~\cite{tangsengwacv18}. Their average score is 98.37 (out of 100).
    \item For each item and its feature $f$ in each base outfit, we create 10 mod samples in the following way:
    \begin{enumerate}[label*=\arabic*]
        \item 500 mod samples are first created by changing the item-feature $f$ in the base sample. 
        In the case of $edge\_image$, we replace it with that of other item occupying the same part of an outfit randomly chosen from the test partition of the dataset. In the case of $colors$, we replace it with random colors. \label{step:replacing_item_feature}
        \item Their scores are computed by the outfit grader and the worst ten samples are selected and all the others are discarded. \label{step:replacing_item_ensure}
    \end{enumerate}
\end{enumerate}
Step~\ref{step:replacing_item_ensure} ensures that the grader gives low scores to the created outfits with a replaced item-feature pair. 
For the two features of \textit{edge\_image} and \textit{colors}, the above procedure produces two datasets, which we call \textit{edge\_image}-wise and \textit{colors}-wise samples, respectively. 
Additionally, we create ``item-wise'' samples by replacing the entire item in Step~\ref{step:replacing_item_feature}. An example of created negative samples is shown in Fig.~\ref{table:outfit_flaw_detection_samples_with_score}.
The statistics of the base samples and the three types of negative  samples
are shown in Table~\ref{table:stats_of_outfit_flaw_detection_samples}. The distributions of scores for these samples are shown in Fig.~\ref{fig:sample_score_distribution}. 

\begin{table}[t]
\small
\centering
\caption{Statistics of the base samples and the negative samples created from them. The three types of negative samples, i.e.,  \textit{edge\_image}-wise, \textit{colors}-wise, and item-wise, have identical statistics by their construction.}
\label{table:stats_of_outfit_flaw_detection_samples}
\begin{tabularx}{\columnwidth}{l *{4}{Y}}
\toprule
Sample type & \multicolumn{4}{c}{Number of samples containing following} \\ 
\cmidrule{2-5} 
 & \multicolumn{2}{c}{outfit parts} & \multicolumn{2}{c}{number of items} \\ 
 \midrule
Base sample & \begin{tabular}[c]{@{}l@{}}Outer\\ Upper\\ Lower\\ Full\\ Feet\\ Accessory0\\ Accessory1\\ Accessory2\end{tabular} & \begin{tabular}[c]{@{}l@{}}205\\ 682\\ 715\\ 330\\ 967\\ 986\\ 901\\ 691\end{tabular} & \begin{tabular}[c]{@{}l@{}}3 items\\ 4 items\\ 5 items\\ 6 items\\ 7 items\\ 8 items\\\midrule \textbf{Total}\end{tabular} & \begin{tabular}[c]{@{}l@{}}14\\ 98\\ 396\\ 383\\ 107\\ 2\\\midrule \textbf{1,000}\end{tabular} \\ \midrule
\begin{tabular}[]{@{}l@{}}Outfit flaw \\ detection sample\end{tabular} & \begin{tabular}[c]{@{}l@{}}Outer\\ Upper\\ Lower\\ Full\\ Feet\\ Accessory0\\ Accessory1\\ Accessory2\end{tabular} & \begin{tabular}[c]{@{}l@{}}2,050\\ 6,820\\ 7,150\\ 3,300\\ 9,670\\ 9,860\\ 9,010\\ 6,910\end{tabular} & \begin{tabular}[c]{@{}l@{}}3 items\\ 4 items\\ 5 items\\ 6 items\\ 7 items\\ 8 items\\\midrule \textbf{Total}\end{tabular} & \begin{tabular}[c]{@{}l@{}}420\\ 3,920\\ 19,800\\ 22,980\\ 7,490\\ 160\\\midrule \textbf{54,770}\end{tabular} \\ \bottomrule
\end{tabularx}
\vspace{-4mm}
\end{table}

\subsection{Results}

\begin{figure}[t]
\scriptsize
\centering

\begin{tabularx}{\columnwidth}{NY}
\toprule
Sample Type & \begin{tabular}[]{@{}c@{}}Items in outfit, its features, and \textit{IFIV} scores of each feature\end{tabular}\\
\midrule
{\begin{tabular}[]{@{}c@{}}base\\sample\end{tabular}} & 
{\begin{tabular}[]{@{}c@{}}
    \includegraphics[width=.7\columnwidth]{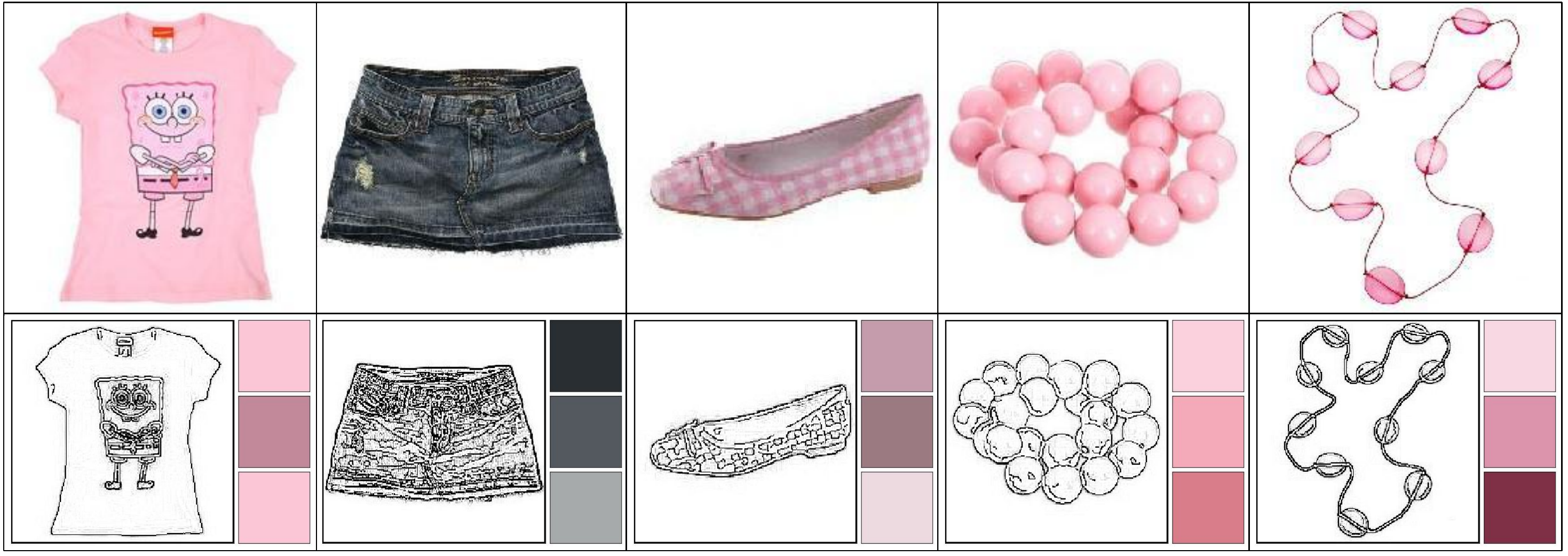}\\
    \includegraphics[width=.7\columnwidth]{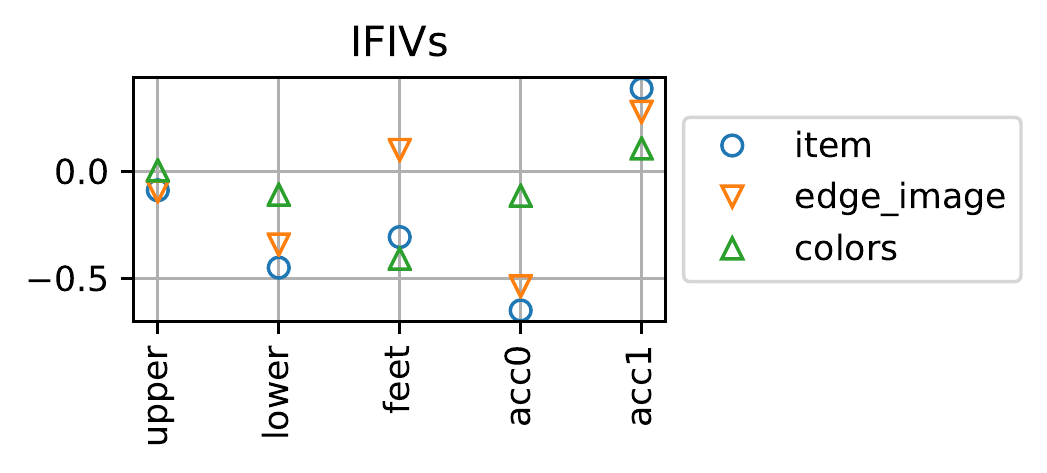}
\end{tabular}} \\
\midrule
{\begin{tabular}[]{@{}c@{}}item-\\wise\end{tabular}} & 
{\begin{tabular}[]{@{}c@{}}
    \includegraphics[width=.7\columnwidth]{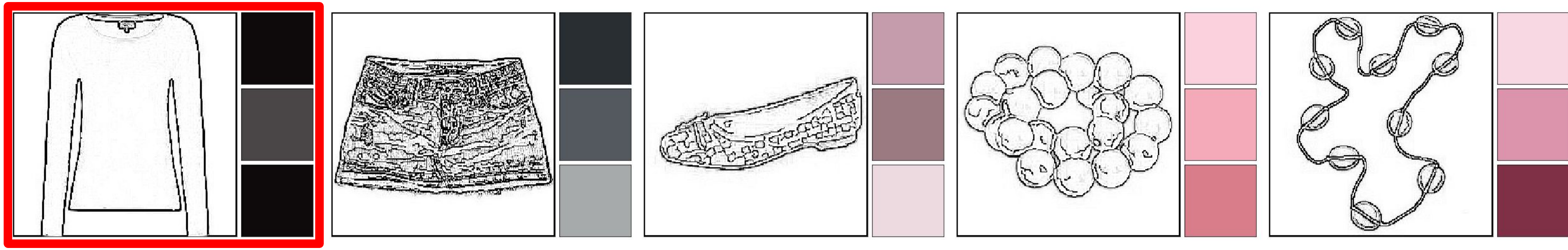}\\
    \includegraphics[width=.7\columnwidth]{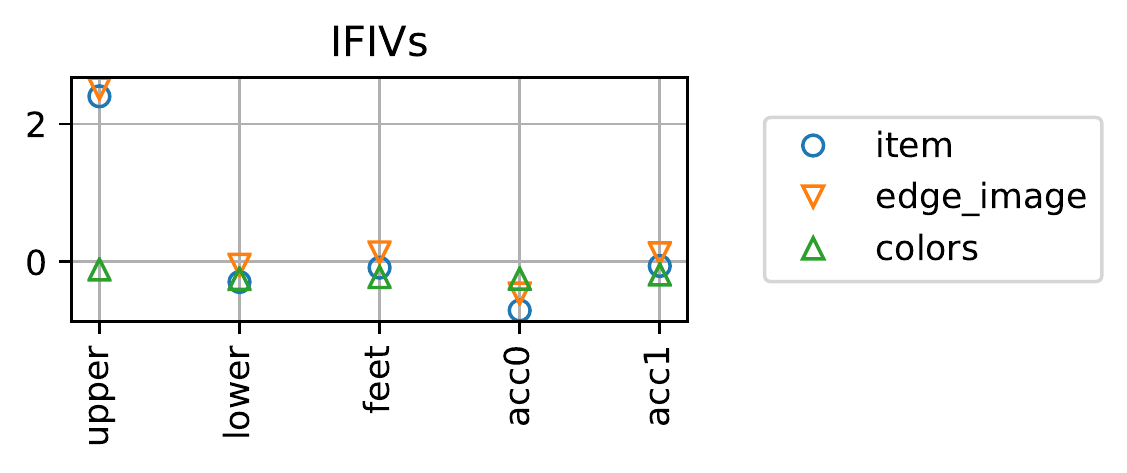}
\end{tabular}} \\
\midrule
{\begin{tabular}[]{@{}c@{}}edge-\\image-\\wise\end{tabular}} & 
{\begin{tabular}[]{@{}c@{}}
    \includegraphics[width=.7\columnwidth]{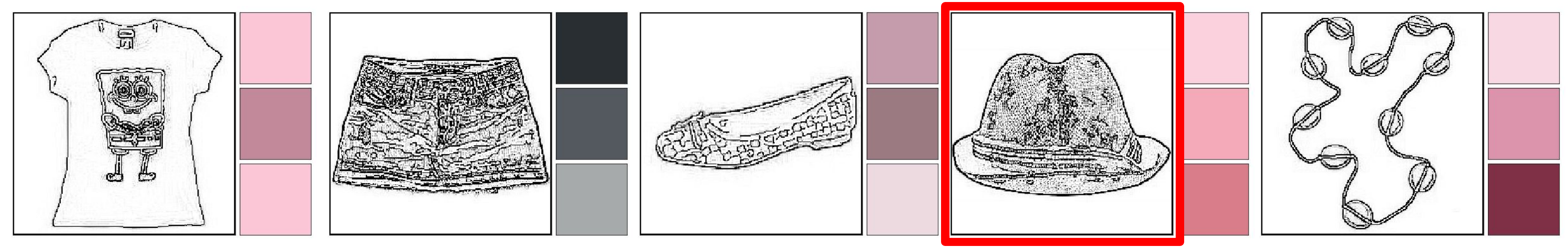}\\
    \includegraphics[width=.7\columnwidth]{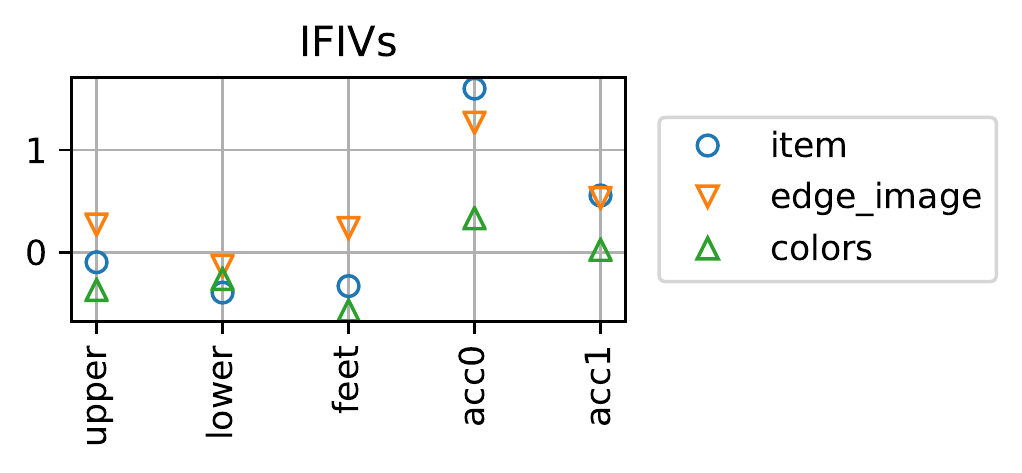}
\end{tabular}} \\
\midrule
{\begin{tabular}[]{@{}c@{}}base\\sample\end{tabular}} & 
{\begin{tabular}[]{@{}c@{}}
    \includegraphics[width=.7\columnwidth]{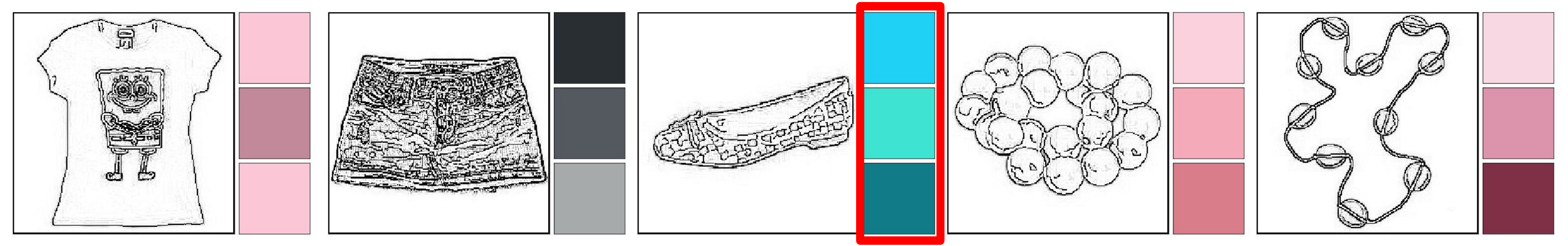}\\
    \includegraphics[width=.7\columnwidth]{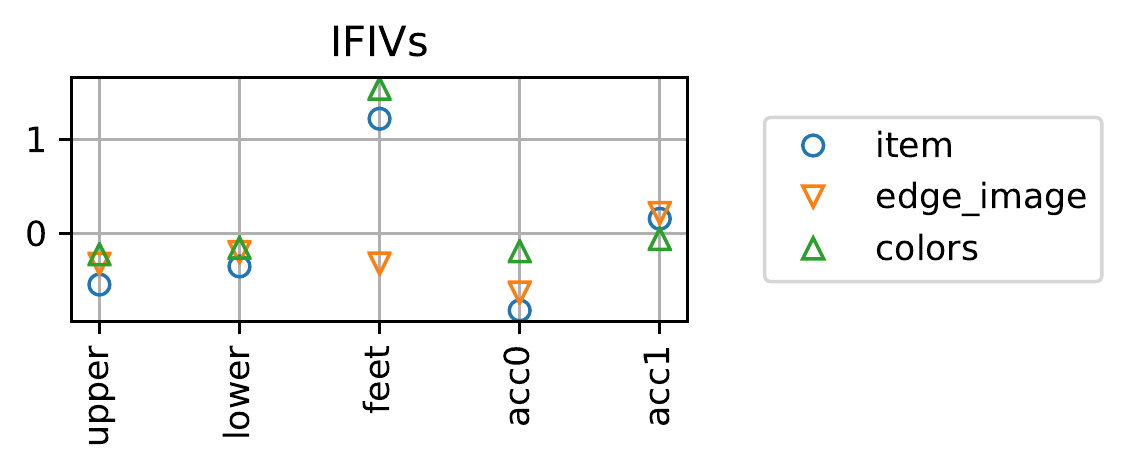}
\end{tabular}} \\
\bottomrule
\end{tabularx}
\vspace{-1em}
\caption{An example of computation of IFIVs. The red boxes indicate the replaced entities from the original high-quality outfits, which makes the new outfits have low outfit scores. ``IFIV score'' means negative IFIV value.
}
\vspace{-1.5em}
\label{table:outfit_flaw_detection_samples_with_score}
\end{figure}

We apply our method to the three types of samples created as explained above. To be specific, inputting each sample to the grader, which yield a lower score as explained above, we compute \textit{IFIV}s for the score defined in (\ref{equation:IFIV}).
We then find the part with the minimum \textit{IFIV}, or equivalently, that the maximum negative \textit{IFIV} 
over all features $f(\in \{edge\_image, colors\})$ as
\begin{align}
    i^* = \argmax _{i,f} (- IFIV_{i,f}).
\end{align}
We regard the prediction 
$i^*$
as correct if it matches the true 
item, which is the replaced one when creating the negative sample. Figure~\ref{table:outfit_flaw_detection_samples_with_score} shows examples of IFIVs for different types of samples. It is seen that the replaced item-feature pairs yield high negative IFIVs, meaning that our method can successfully detect the item lowering the outfit score with the reason why it is bad (i.e., the feature lowering the outfit score).

Table~\ref{table:overall_accuracy_outfit_flaw_detection} show the performance over all the samples. The proposed method can detect the replaced items for \textit{item-}wise samples with 99.51\% accuracy and those for \textit{edge\_image}-wise samples with 98.99\% accuracy, respectively. The accuracy for \textit{colors}-wise samples is 81.83\% and is lower than the others. This is due to the fact that the scores of the \textit{colors}-wise samples tend to be higher and their gap to the original outfits are smaller than the other two types, as shown in Fig.~\ref{fig:sample_score_distribution}. That said, this is fairly good considering the chance rate. Note that for the samples of \textit{edge\_image}-wise and \textit{colors}-wise, it is necessary to predict both the feature and the item correctly.


Table~\ref{table:accuracy_outfit_flaw_detection_by_n_items} shows  accuracy values for different numbers of items. They are quite consistent for \textit{item}- and \textit{edge\_image}-wise samples, except for the outfit with eight items. Note that there is only two out of 1,000 base samples that has eight items, as shown in Table~\ref{table:stats_of_outfit_flaw_detection_samples}, and thus the performance for eight items could be statistically unreliable. For \textit{colors}-wise samples, there is a tendency that the accuracy decreases as the number of items increases.

\begin{table}[t]
\centering
\caption{Overall accuracy (\%) of detection of replaced item-feature pairs.}
\label{table:overall_accuracy_outfit_flaw_detection}
\small 
\begin{tabularx}{\columnwidth}{l *{2}{Y}}
\toprule
Method & Sample type & Prediction accuracy \\ 
\midrule
\begin{tabular}[]{@{}l@{}}Random\end{tabular} & 
\begin{tabular}[]{@{}c@{}}item-wise\\feature-wise\end{tabular} & \begin{tabular}[]{@{}c@{}}18.26\\9.13\end{tabular}  \\ 
\midrule
\begin{tabular}[]{@{}l@{}}Proposed\\method\end{tabular} & \begin{tabular}[]{@{}c@{}}item-wise\\\textit{edge\_image}-wise\\\textit{colors}-wise\end{tabular} & \begin{tabular}[]{@{}c@{}}99.51 \\98.99 \\81.83\end{tabular} \\ 
\bottomrule
\end{tabularx}
\end{table}

\begin{table}[t]
\centering
\caption{Accuracy (\%) of replaced item-feature detection for different numbers of items contained in each outfit. The \textit{By chance} column shows the chance rate for  feature-wise samples.}
\label{table:accuracy_outfit_flaw_detection_by_n_items}
\small
\begin{tabular}{@{}ccccc@{}}
\toprule
\multirow{2}{*}{\begin{tabular}[c]{@{}c@{}}Number \\ of items\end{tabular}} & \multirow{2}{*}{\begin{tabular}[c]{@{}c@{}}By \\ chance\end{tabular}} & \multicolumn{3}{c}{Proposed method (by sample type)} \\ \cmidrule{3-5} 
 &  & item & \textit{edge\_image} & \textit{colors} \\ \midrule
3 & 16.67 & 95.71 & 95.71 & 76.43 \\ 
4 & 12.50 & 99.90 & 97.37 & 86.91 \\ 
5 & 10.00 & 99.72 & 98.94 & 85.39 \\ 
6 & 8.34 & 99.51 & 99.26 & 79.57 \\ 
7 & 7.15 & 99.39 & 99.57 & 76.92 \\ 
8 & 6.25 & 80.00 & 86.25 & 86.25 \\ \bottomrule
\end{tabular}
\end{table}

\begin{table}[t]
\small
\centering
\caption{Accuracy (\%) of replaced item-feature detection classified by different outfit parts. Note that there are eight outfit parts in Polyvore409k dataset; the \textit{By chance} column shows the chance rate for feature-wise samples.}
\label{table:accuracy_outfit_flaw_detection_by_part}
\begin{tabular}{@{}ccccc@{}}
\toprule
\multirow{2}{*}{\begin{tabular}[c]{@{}c@{}}Outfit\\ part\end{tabular}} & \multirow{2}{*}{\begin{tabular}[c]{@{}c@{}}By \\ chance\end{tabular}} & \multicolumn{3}{c}{Proposed method (by sample type)} \\ \cmidrule(l){3-5} 
 &  & item & \textit{edge\_image} & \textit{colors} \\ \midrule
outer & 7.77 & 100.00 & 99.66 & 58.93 \\ 
upper & 8.58 & 99.75 & 99.96 & 57.95 \\ 
lower & 8.59 & 99.40 & 99.36 & 68.20 \\ 
full & 9.93 & 96.36 & 87.70 & 66.36 \\ 
feet & 8.87 & 99.65 & 99.38 & 90.91 \\ 
accessory0 & 8.88 & 99.68 & 99.69 & 94.07 \\ 
accessory1 & 8.72 & 99.76 & 99.99 & 89.39 \\ 
accessory2 & 8.49 & 100.00 & 99.99 & 93.70 \\ \bottomrule
\end{tabular}
\end{table}

Table~\ref{table:accuracy_outfit_flaw_detection_by_part} shows accuracy values calculated for each part of outfits. It is seen that  for \textit{item}- and \textit{edge\_image}-wise samples, the performance are almost the same across all outfit parts, except the full outfit part showing slightly lower accuracy. For \textit{colors}-wise samples, the accuracies are lower the other two types and are somewhat different for different parts. 

\section{Conclusion}\label{sec:ch4_conclusion}
In this paper, we have proposed a novel method for item-feature-wise explanation of outfits. The method can quantify the effect of interpretable features of each item on the goodness of an outfit with the proposed \textit{Item Feature Influence Value (IFIV)}. It does not need any item-level attribute annotation. Using the \textit{IFIV} of each item-feature pair in an outfit, we can detect the bad item in an outfit lowering its score by finding the item-feature pair with the maximum negative \textit{IFIV}. The experiments have shown that our method can detect the bad items at 99.51, 98.99, and 81.83\%, for datasets of item-wise, \textit{edge\_image}-wise, and \textit{colors}-wise samples, respectively. 

{
\bibliographystyle{ieee}
\bibliography{my}
}

\end{document}